\documentclass{article} 
\usepackage[final]{ACL2023}
\usepackage{times}  
\usepackage{helvet}  
\usepackage{courier}  
\usepackage{graphicx} 
\usepackage{natbib}  
\usepackage{caption} 
\usepackage{algorithm}
\usepackage{svg}
\usepackage{listings}
\lstdefinelanguage{MyDiff}{
  morecomment=[f][\color{gray}]{@@},
  morecomment=[f][\color{blue}]{authors_comments},
  morecomment=[f][\color{blue}]{\\},
  morecomment=[f][\color{gray}]{diff},
  morecomment=[f][\color{gray}]{index},
  morecomment=[f][\color{gray}]{---},
  morecomment=[f][\color{gray}]{+++},
  morecomment=[f][\color{red}]{-},
  morecomment=[f][\color{green!70!black}]{+},
  morestring=[b]",
        literate=
      {ℝ}{{$\mathbb{R}$}}1
      {→}{{$\to$}}1
      {∀}{{$\forall$}}1
      {₀}{{$_0$}}1
      {∨}{{$\vee$}}1
}

\lstdefinelanguage{prompts}{
  moredelim=**[is][\normalfont]{@}{@}, 
}

\lstnewenvironment{promptlisting}[1][]{
  \lstset{
    language=prompts,
    basicstyle=\ttfamily\small,
    breaklines=true,
    breakatwhitespace=false,
    breakindent=1em,
    frame=single,
    tabsize=2,
    showstringspaces=false,
    captionpos=b,
    xleftmargin=0pt,
    xrightmargin=0pt,
    linewidth=\linewidth, 
    postbreak=,
    #1
  }
}{}

\usepackage[english]{babel}
\usepackage{xcolor}
\usepackage{soul}
\usepackage{amsmath}
\usepackage{adjustbox}
\usepackage{booktabs}
\usepackage{subcaption}
\usepackage{multirow}
\usepackage{tikz}
\usepackage[noend]{algpseudocode}
\usepackage[inline]{enumitem}
\usepackage[colorinlistoftodos,prependcaption,textsize=tiny]{todonotes}
\usetikzlibrary{fit, positioning, backgrounds, decorations.pathreplacing, calc, arrows}
\usepackage{enumitem}
\usepackage{tcolorbox}
\usepackage{verbatim}
\usepackage{ragged2e}
\usepackage{amssymb}
\usepackage{amsmath}
\usepackage[framemethod=TikZ]{mdframed}
\usepackage{wrapfig}
\newmdenv[%
    backgroundcolor=gray!10,
    linecolor=black,
    outerlinewidth=0.5pt,
    roundcorner=1mm,
    skipabove=\topsep,
    skipbelow=\topsep,
    font=\ttfamily\tiny,
]{promptbox}
\newcommand{\ignore}[1]{}

\newcommand{\gustavo}[1]{\textcolor{brown}{\textbf{GS:} #1}}
\newcommand{\priyanshu}[1]{\textcolor{blue}{\textbf{PG:} #1}}

\newcommand{\method}[0]{$\textsc{STACKFEED}$}
\newcommand{\promptAgent}[0]{$\textsc{PromptAgent}$}
\newcommand{\embeddingModel}[0]{\textsc{OpenAI-Text-Embedding-3-Large}}

\newcommand{\methodfull}[0]{\textbf{S}tructured \textbf{T}extual \textbf{A}ctor-\textbf{C}ritic \textbf{K}nowledge base editing with \textbf{FEED}back}

\usepackage{amsmath,amsfonts,bm}

\def\eqref#1{equation~\ref{#1}}

\def\1{\bm{1}}

\DeclareMathAlphabet{\mathsfit}{\encodingdefault}{\sfdefault}{m}{sl}
\SetMathAlphabet{\mathsfit}{bold}{\encodingdefault}{\sfdefault}{bx}{n}

\usepackage{wrapfig}
\usepackage{hyperref}
\usepackage[utf8]{inputenc}
\usetikzlibrary{chains, positioning, calc, shapes}
\newcommand*\circled[1]{\tikz[baseline=(char.base)]{
            \node[shape=circle,draw,inner sep=1pt] (char) {\small #1};}}

\title{\method: \methodfull}

\author{
{Shashank Kirtania\thanks{\quad Equal contribution.}\;\textsuperscript{~1} \quad   Naman Gupta\footnotemark[1]\; \textsuperscript{2} \quad  Priyanshu Gupta\textsuperscript{1}} \\ ~\textbf{Sumit Gulwani\textsuperscript{3}} \quad \textbf{Arun Iyer\textsuperscript{2}}\quad \textbf{Suresh Parthasarathy\textsuperscript{2}}\quad \textbf{Arjun Radhakrishna\textsuperscript{3}}\\ ~\textbf{Sriram K. Rajamani\textsuperscript{2}} \quad \textbf{Gustavo Soares\textsuperscript{3}} \\
    \textsuperscript{1}Microsoft, Bengaluru
  \textsuperscript{2}Microsoft Research India, Bengaluru 
  \textsuperscript{3}Microsoft, Redmond \\
  \texttt{\{ t-shkirtania, t-nagupta, priyansgupta\}@microsoft.com}\\
  \texttt{\{sumitg, ariy, supartha, arradha, sriram, gsoares\}@microsoft.com
  }
}

\begin{document}

\maketitle
\begin{abstract}
Large Language Models (LLMs) often generate incorrect or outdated information, especially in low-resource settings or when dealing with private data. To address this, Retrieval-Augmented Generation (RAG) uses external knowledge bases (KBs), but these can also suffer from inaccuracies. We introduce \method, a novel \methodfull~approach that iteratively refines the KB based on expert feedback using a multi-actor, centralized critic reinforcement learning framework. \method~defines a ReACT actor agent on each document to perform structured edits based on document-specific targeted instructions.  Experimental results showcase that \method~significantly improves KB quality and performance of the RAG system. We evaluate \method~on low-resource programming problems, modified python packaged and factual question-answering tasks. 
\end{abstract}
\section{Introduction}
\label{sec:introduction}
Large Language Models (LLMs) often produce incorrect or outdated information, particularly in low-resource settings or when handling private data. Even if the information provided is accurate, LLMs can generate hallucinated or imaginary content alongside it~\citep{10.1145/3728894, maynez-etal-2020-faithfulness, zhou-etal-2021-detecting}\ignore{\gustavo{I did not understand this sentence. The information is correct but it hallucinates?}}. A promising solution to address these issues is the integration of retrieval components that extract relevant information from external knowledge sources, known as Retrieval-Augmented Generation (RAG)~\citep{drqa,atlas,replug, li2025surveypersonalizationragagent}. For clarity, we will refer to these external knowledge sources as Knowledge Bases (KBs). However, KBs themselves can suffer from inaccuracies, incompleteness, or outdated content. To address these challenges, there is growing interest in Knowledge Editing (KE) techniques to enhance LLMs with up-to-date and accurate knowledge.

Advancements in KE have focused on updating the model’s parameters~\citep{cl2,rome,memit}, adding new parameters to model~\citep{transformerpatcher, melo}, and holding additional memory~\citep{memprompt,ike,remake}. Contrary to approaches that update model parameters or add new parameters that require white-box access to LLMs, memory-based approaches can work with black-box access to LLMs. In a similar line of thought, recently, KE approaches have also focused on refining the KBs themselves~\citep{erase}. For example, the method proposed by~\citet{erase} continuously updates KBs with new information when presented with a document containing the exact information to be updated, such as updating the current identity of the British Prime Minister in the KB when the news of election results is provided. This approach demonstrates that directly editing the KB is more effective than simply adding new documents, which may coexist with outdated or inaccurate ones. Removing older documents is often non-trivial, as only certain sections may be incorrect, while other parts could still provide valuable information for different queries. 

However, in applications such as chatbots or code generation tools that rely on API documentation, up-to-date information may not always be readily available in structured documents~\citep{cataractbot,ragchatbot,liu2025codeupdatearenabenchmarkingknowledgeediting}. In these scenarios, expert feedback becomes essential—not only for correcting erroneous outputs from the LLM but also for directly updating the underlying knowledge base (KB) with accurate information. This need is particularly pressing in live systems that depend on real-time, reliable data. Domains like healthcare~\citep{cataractbot}, legal tech~\citep{liu2025codeupdatearenabenchmarkingknowledgeediting}, and financial services demand high precision and instant updates. Ensuring continuous and trustworthy KB revisions is therefore critical to maintaining the safety, effectiveness, and reliability of RAG applications in such high-stakes environments.

\begin{figure*}
\newsavebox\myv
\begin{lrbox}{\myv}
\begin{tikzpicture}[
  every node/.style={rounded corners}
]

\node [text width=0.4\textwidth, fill=orange!30] (task) {
{\tiny
\noindent\textbf{Task:}
    Given an array of integers nums, write a function that returns the number of
    good pairs. A pair \texttt{(i, j)} is called good if \texttt{nums[i] ==
    nums[j]} and \texttt{i < j}.
\vspace{2ex}

\noindent\textbf{Retrieved Documents:}
\vspace{-1.5ex}
        \begin{verbatim}
    builtin-array.md
    collections-persistent-vec.md
    math-fibonacci.md
    random-dice.md
        \end{verbatim}
\vspace{2ex}
        
\noindent\textbf{Output Program:}
\vspace{-1.5ex}
    \begin{verbatim}
fun numIdenticalPairs(ns: Array[I32]): I32 =>
  var count: I32 = 0
  for i in Range(0, ns.size() - 1) do
    for j in Range(i + 1, ns.size()) do
      if ns(i) == ns(j) then
        count = count + 1
        ...
    \end{verbatim}
}};
\draw [decorate, decoration={brace,amplitude=2mm,raise=2ex,mirror}, thick]
    (task.north west) -- (task.south west) 
    node[midway,  anchor=east, xshift=-2.5mm] {{\tiny\tabular{c}Simulate\\Task\endtabular}}
    ;

\node [
    text width=0.5\textwidth,
    right=of task.north east,
    anchor=north west,
    xshift=0.2\textwidth,
    fill=pink!30
] (kb) {
{\tiny \textbf{Knowledge Base}
\\
\vspace{-1.5ex}
\begin{verbatim}
    builtin-array.md
    collections-persistent-vec.md
    math-fibonacci.md
    ...
\end{verbatim}
}};

\draw [decorate, decoration={brace,amplitude=2mm,raise=-1.2cm}]
    ($(task.north east)!0.2!(task.south east)$)
    --
    ($(task.north east)!0.47!(task.south east)$)
    node[midway, anchor=west, xshift=-4mm] (retrieval) {{\tiny \textbf{Retrieval}}}
    ;
\path [draw]
    (kb.west) edge [out=180,in=20,->] (retrieval.east) ;
\path [draw]
    (retrieval.west) edge[->]  ($(retrieval.west) + (-0.5cm,0)$) ;

\node [
    text width=0.5\textwidth,
    below=of kb.south west,
    anchor=north west,
    fill=blue!30,
    yshift=0.5cm
] (errors) {
{\tiny
\noindent\textbf{Expert (Compiler) Feedback}
    \begin{verbatim}
main.pony:15:16: call is not partial but the method is
  - a question mark is required after this call
      if nums(i) == nums(j) then
               ^
main.pony:15:27: call is not partial but the method is
  - a question mark is required after this call
      if nums(i) == nums(j) then
                          ^
main.pony:11:42: function signature is not marked as
  partial but the function body can raise an error
      fun numIdenticalPairs(nums: Array[I32]): I32 =>
                                                  ^
    \end{verbatim}
}};
\draw [decorate, decoration={brace,amplitude=2mm,raise=0.1cm}]
    ($(task.north east)!0.60!(task.south east)$)
    --
    ($(task.north east)!0.95!(task.south east)$)
    node[midway, anchor=west, xshift=5mm] (feedback) {{\tiny \textbf{Feedback}}}
    ;
\path [draw]
    (errors.west) edge [out=180,in=20,<-] (feedback.east) ;
\path [draw]
    (feedback.west) edge[<-]  ($(feedback.west) + (-0.2cm,0)$) ;

\node [
    text width=0.6\textwidth,
    below=of task.south west,
    anchor=north west,
    fill=red!10,
    xshift=-0.4cm,
    yshift=-0.6cm,
    minimum height=3cm,
    draw,
    dotted,
    on background layer
] (reflection-3) { };
\node [
    text width=0.6\textwidth,
    below=of task.south west,
    anchor=north west,
    fill=red!20,
    xshift=-0.7cm,
    yshift=-0.3cm,
    minimum height=3cm,
    draw,
    dotted,
    on background layer
] (reflection-2) { };
\node [
    text width=0.6\textwidth,
    below=of task.south west,
    anchor=north west,
    fill=red!30,
    xshift=-1cm
] (reflection) {
{\tiny
\noindent\textbf{Reflection:}
The output code contains the following errors:

1. The array access calls \texttt{nums(i)} and \texttt{nums(j)} are not marked as partial, but
they should be as the apply function in the Array class is partial and can throw errors.
The corrected code is: \texttt{nums(i)? = nums(j)?}
To further illustrate the fix, consider the following simplified example:\\
Code with error:\\
\texttt{~~~fun get(arr: Array[I32], index: USize): I32 ? => arr(index)}\\
Fixed code:\\
\texttt{~~~fun get(arr: Array[I32], index: USize): I32 ? => arr(index)?}\\
...

\noindent\textbf{Selection:} \texttt{builtin-array.md}\\
\noindent\textbf{Reasoning:}
This file contains information about the Array class, including apply...
}};

\node[
    draw,
    dashed,
    fit={(reflection) (reflection-2) (reflection-3)},
    inner sep=4mm,
    yshift=-3mm
] (collected-reflections) { };

\node[below=of reflection.south, yshift=4mm] {
{\small
    Collected selection reasoning for \texttt{builtin-array.md} 
}};

\path [draw]
    ($(errors.west)!0.70!(errors.south west)$)
    edge [->,>=stealth',out=230,in=30]
    node[fill=white,yshift=0.5mm] {{\tiny \textbf{Centralized Critic}}}
    (reflection.north) ;

\node [
    text width=0.6\textwidth,
    below=of collected-reflections.south,
    anchor=north,
    fill=yellow!30,
] (gradient) {
{\tiny
\noindent\textbf{Document-wise Partial Gradient:}
\texttt{builtin-array.md} is a KB document detailing
the use of the methods and functions of the \texttt{Array} class...
\\
~\\
Method Availability and Error Handling: 
The current file lacks explicit guidance on the partiality of methods and
the use of~\texttt{?} operator. The omission can lead developers to...
The file should accurately reflect the partiality of methods like \texttt{apply}
and \texttt{update}...\\
}};

\path [draw]
    (collected-reflections.south)
    edge [->]
    node[fill=white,yshift=0.5mm] {{\tiny \textbf{Aggregate Document Gradients}}}
    (gradient.north) ;

\node [
    text width=0.4\textwidth,
    right=of gradient.south east,
    anchor=south west,
    fill=green!30,
    xshift=2cm,
    yshift=1cm
] (edit) {
{\tiny
\noindent\textbf{Edit:} Add Section
\\
When working with arrays in Pony, it is important to understand that
some methods are partial, meaning the may not  succeed in all cases and might
raise an error. To safely handle these errors, Pony provides the ? operator...
\\
The \textbf{apply} method on arrays is partial and will raise an error in the
index is out of bounds...
\\
For example, the `apply` method on arrays is partial and requires error handling:

```pony
\\
try
\\
  my\_array.apply(42)? == 7
\\
end
\\
```
\\
In this snippet, if the index 42 is out of bounds
\\
the `apply` method will raise an error, which is then handled by the `try` block, preventing the program from crashing.
}};

\path [draw]
    (gradient.east)
    edge [->, in=180, out=0]
    node[fill=white,yshift=0.5mm] {{\tiny \textbf{Per-Document Actor}}}
    (edit.west) ;
\end{tikzpicture}
\end{lrbox}

\resizebox{\textwidth}{!}{\usebox\myv}
\caption{Example of the \method\ pipeline in the ARKS Pony scenario. We explain the example in more detail in appendix \ref{sec:exampleoverview}}
\label{fig:flowchart}
\end{figure*}
To leverage expert or oracle feedback, we propose \method, a \methodfull~technique. Our contributions are as follows: \begin{enumerate}
    \item \textbf{Introduction of Feedback-Driven KB Editing}: We present a novel framework that refines the knowledge base using structured edits based on oracle or expert feedback.
    \item \textbf{Definition and Evaluation of KB Characteristics}: We define desirable characteristics for knowledge base refinement, including coherence, completeness and introduce corresponding metrics to quantitatively assess these properties. 
    
    \item \textbf{Empirical Evaluation and Performance Gains}: We demonstrate that \method~significantly improves the accuracy and reliability of RAG system in a variety of settings. 
\end{enumerate} 

\section{Related work}
\label{sec:related}
The \method~framework addresses a key limitation of current RAG systems: the inability to dynamically update Knowledge Bases (KBs) without retraining or altering model parameters. Our work draws from research in Retrieval-Augmented Generation (RAG), Continual Learning and incorporating insights from Multi-Agent Reinforcement Learning (MARL) to propose an effective solution for KB editing.

\textbf{Retrieval Augmented Generation (RAG):} RAG systems enhance LMs by retrieving relevant knowledge from a KB based on the input query and appending it to the context, thereby addressing the limitations of standalone LMs that lack sufficient context and produce inaccurate answers~\citep{drqa,knnlm,realm,atlas,replug}. These systems dynamically construct contexts from unstructured KBs without modifying the LM’s internal parameters. \method~further enhances RAG systems by refining the KB itself based on feedback, ensuring more accurate and up-to-date information.
Recent works showcase the failure of RAG due to inconsistencies in the retrieved documents \cite{xiang2024certifiablyrobustragretrieval, wang2025astuteragovercomingimperfect}. 

\textbf{Knowledge Editing:} Knowledge Editing approaches fall into two categories~\citep{yao-etal-2023-editing}: \textbf{Model Editing}, which modifies the LM parameters directly, and \textbf{Memory Based}, which updates the knowledge supplied to the model. While Model Editing efficiently alters specific facts using specialized secondary models or altering parameters~\citep{knowledgeeditor,memit}, it struggles to ensure consistent updates across contexts~\citep{kechallenge,recoe}. In contrast, Memory Based methods either add an auxiliary counterfactual model trained on new information~\citep{mitchell2022memory} or they modify the KB itself, enabling updates to be reflected in outputs without changing model parameters~\citep{remake,erase}. \method~builds on memory based techniques by leveraging expert feedback to refine the KB systematically, ensuring more accurate and consistent responses.

\textbf{Prompt Optimization:} With the advent of LMs, some recent works approximate gradients in text-based environments using LMs~\citep{protegi,wang2023promptagent,kirtania-etal-2024-logic,gupta2024metareflection} for optimizing task prompts. \method~is inspired by these approaches and generates textual reflections, similar to MetaReflection~\citep{gupta2024metareflection} and \cite{refelction}, as proxies for gradients. It provides actionable guidance for document updates without the need for differentiable models. Additionally, \method~adopts clustering strategies for feedback aggregation from works like UniPrompt \citep{taskfacetstructured}- ensuring that actors receive coherent and non-redundant instructions.
\begin{figure*}[htbp]
    \centering
    \includegraphics[width=0.97\textwidth]{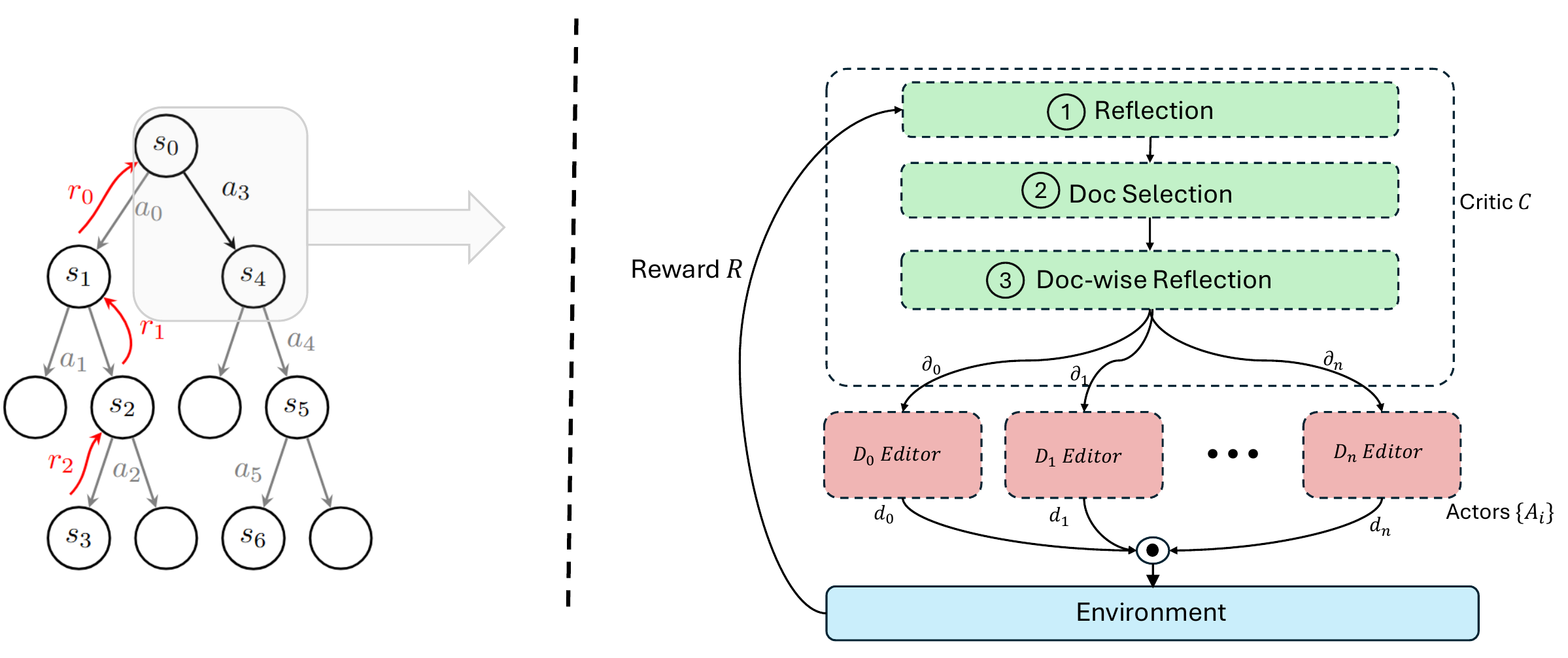} 
\caption{a) MCTS (Monte Carlo Tree Search) planning for state search. The tree structure enables strategic planning for \method. b) A simplified state transition example. Upon receiving a reward from the environment (or expert) on the given state of the knowledge base (KB) \textcircled{$s_0$}, a centralized critic \textcircled{1} generates a reflection on observed failures to calculate the textual gradient. The critic uses this reflection to select documents responsible for the error and \textcircled{2} assigns credit to actors in the form of document-wise reflections. The actors then iteratively edit the documents to reach state \textcircled{$s_4$}.}
    \label{fig:example}
\end{figure*}
\section{Problem Formulation}

Typical RAG systems assume that the information present in the retrieved documents is correct and consistent. Our work focuses on scenarios where incorrect answers are generated due to issues in the retrieved documents from a Knowledge Base (\( \mathcal{K} \)). 

More formally, we define the \( \mathcal{K} \) as a collection of documents \( D_i \) for \( i = 1, \ldots, N \). Each document \( D_i \) can be represented as a set of chunks \( c_{ij} \). The state of $\mathcal{K}$ is the specific configuration of all chunks within it.

For a given query \( q \), such as the code generation Task in Figure~\ref{fig:flowchart}, a retriever fetches a set of relevant documents \( \Gamma(q, \mathcal{K}) \). In the example, this corresponds to the Retrieved Documents list, which includes \texttt{builtin-array.md} and \texttt{collections-persistent-vec.md}.

An LLM, \( M \), then generates a response \( r \) based on the query and the retrieved documents, i.e., \( r = M(q, \Gamma(q, \mathcal{K})) \). The initial Output Program in Figure~\ref{fig:flowchart} is an instance of such a response.

However, this response \( r \) may be incorrect. We obtain feedback on the correctness of \( r \), for instance, from Expert (Compiler) Feedback as shown in Figure~\ref{fig:flowchart}. This feedback reveals flaws in the generated program that stem from deficiencies in \( \mathcal{K} \). The expert is used as a scoring function, \( g \) which evaluates whether a response \( r \) is correct or not.


Our goal is to optimize the Knowledge Base state \( \mathcal{K} \) to state \( \mathcal{K} \)* which maximizes the scores for the responses for a batch of queries \( \mathcal{Q} \) by learning from expert feedback:
\begin{gather}
    \mathcal{K}^* = \arg \max_{\mathcal{K}} \frac{1}{|\mathcal{Q}|} \sum_{q_i \in \mathcal{Q}} g(\mathcal{M}(q_i, \Gamma(q_i, \mathcal{K})))
\label{eq:optim_objective}
\end{gather}

\section{Methodology}

We propose \method, an agent that employs Monte Carlo Tree Search (MCTS) to search for an optimal state of the knowledge base (KB). \method~utilizes a \textit{multi-actor}, \textit{centralized critic} reinforcement learning architecture to guide transitions between KB states. This setup enables efficient exploration of a large, structured edit space \cite{wang2023promptagent, gupta2024metareflection}, facilitating strategic and interpretable knowledge refinement.

The use of a multi-actor architecture with a centralized critic is central to \method's design. In this formulation, each document within the KB is managed by a dedicated actor responsible for localized edits, while a centralized critic provides joint feedback by aggregating signals across all actor-document interactions. Our design aligns with recent findings from \citet{lyu2024centralized}, who show that \textit{history–state centralized critics}—critics conditioned on both joint observation histories and global state—can offer accurate and stable policy gradients, particularly in settings with partial observability and distributed decision-making.

In our context, where agents must coordinate to revise distinct portions of a shared KB based on sparse oracle feedback, this centralized view enables effective credit assignment. Specifically, we incorporate mechanisms inspired by counterfactual multi-agent policy gradients (COMA) \citep{foerster2018coma} to attribute responsibility for errors to specific documents and actors. The centralized critic leverages these credit signals to generate high-quality textual reflections that guide each actor's editing trajectory. This design allows \method~to maintain coherence across KB documents while iteratively improving task-specific correctness and completeness.

By combining decentralized editing with centralized, feedback-driven evaluation, \method~learns to make targeted and interpretable updates to the KB. This architecture is particularly suited for real-world retrieval-augmented generation (RAG) settings, where maintaining consistency and relevance across independently edited documents is critical for downstream performance.

\subsection{Knowledge base editing as state search}
We model knowledge base (KB) editing as a state optimization problem over the configuration of documents in the KB. Given a query and retrieved evidence, a language model generates a response. When errors arise due to incomplete or inaccurate evidence, we update the KB such that future responses are more accurate.

In this formulation, each KB state reflects a specific configuration of document contents, and actions correspond to edits applied to these documents. A transition function updates the KB by applying these edits, and a reward function evaluates the quality of the resulting KB state based on how well it supports correct and complete responses across a set of queries.

Our objective is to find the optimal KB state that maximizes this reward. This enables a feedback-driven editing process where the system learns to apply targeted modifications to improve RAG performance in a data-driven and interpretable manner. We define the action space, search space and the optimization objective more formally in the appendix section \ref{sec:KB_as_state_search}.

\label{sec:methodology}

\subsection{Knowledge Base Editing Agent}
We define KB editing agent that operates on a reward signal as a model's performance over a batch of queries for the given knowledge base in a RAG system. 

\textbf{Centralized Critic:}\quad The centralized critic $C$ evaluates the RAG system's performance by analyzing expert feedback and the current knowledge base state. When errors occur, the critic identifies which specific documents caused the problems and generates targeted feedback for improvement.


The critic examines each failed query and its corresponding expert feedback, first reflecting upon it to fully understand the error and then identifying which documents are responsible for the error. Following established methods from prior work~\citep{protegi,taskfacetstructured, gupta2024metareflection}, rather than simply listing all issues in each retrieved document, the critic clusters similar problems together to identify common patterns and generate more generalizable insights.

These aggregated reflections are analogs to partial gradients $\partial_j$  for each document that guide each document-specific actor $A_j$ on improving their assigned documents. This approach ensures document updates address systematic issues rather than isolated errors, leading to more effective knowledge base refinement. By analyzing failures across multiple queries and clustering similar issues, the critic provides more strategic guidance than treating each error in isolation.

\begin{table*}[t]
\centering
\resizebox{\textwidth}{!}{%
\begin{tabular}{@{}ll cc cc cc cc@{}}
\toprule
\multicolumn{2}{c}{} & \multicolumn{2}{c}{\textbf{Pony}} & \multicolumn{2}{c}{\textbf{SciPy}} & \multicolumn{2}{c}{\textbf{Tensorflow}} & \multicolumn{2}{c}{\textbf{CLARK-news}} \\
\cmidrule(lr){3-4} \cmidrule(lr){5-6} \cmidrule(lr){7-8} \cmidrule(lr){9-10}
Model & Method & Acc & $\sigma$ & Acc & $\sigma$ & Acc & $\sigma$ & Acc & $\sigma$ \\
\midrule
\multirow{3}{*}{\textbf{GPT-4Turbo}} & Base KB & 29.99 & 1.57 & 52.04 & 0.00 & 28.88 & 2.18 & 26.27 & 1.20 \\
                                     & \promptAgent-E & 32.22 & 1.57 & 53.40 & 3.12 & 47.77 & 3.57 & 28.80 & 2.39 \\
                                     & \method & \textbf{37.04} & 1.28 & \textbf{59.38} & 1.22 & \textbf{53.84} & 3.11 & \textbf{37.28} & 1.69 \\
\midrule
\multirow{3}{*}{\textbf{GPT-4o}}     & Base KB & 31.41 & 1.28 & 54.13 & 1.22 & 31.75 & 2.91 & 28.80 & 1.69 \\
                                     & \promptAgent-E & 34.21 & 1.49 & 55.27 & 3.05 & 49.03 & 3.62 & 30.01 & 2.41 \\
                                     & \method & \textbf{42.32} & 2.11 & \textbf{61.60} & 2.43 & \textbf{55.32} & 2.18 & \textbf{40.40} & 1.63 \\
\midrule
\multirow{3}{*}{\textbf{GPT-4.1}}    & Base KB & 35.40 & 2.52 & 53.40 & 2.43 & 34.60 & 3.11 & 30.89 & 1.20 \\
                                     & \promptAgent-E & 36.10 & 1.73 & 56.02 & 3.28 & 50.27 & 3.44 & 31.33 & 2.19 \\
                                     & \method & \textbf{45.62} & 3.67 & \textbf{60.83} & 2.84 & \textbf{57.61} & 2.18 & \textbf{43.03} & 2.14 \\
\bottomrule
\end{tabular}%
}
\caption{Correctness performance comparison between \method~and baseline method across multiple datasets, reported as accuracy percentages (higher is better). Best results for each model and dataset are highlighted in bold.}
\label{tab:baseline-comparision}
\end{table*}

\textbf{Actors:}\quad 
Each document \( D_i \in \mathcal{K} \) is managed by a distinct actor, \( A_i \), which is modeled as a ReACT agent \cite{react} responsible for making structured edits to its document. Each actor operates independently, receiving reflections from the centralized critic on how to modify the content of \( D_i = [c_{ij}]\). The actors need to only update these chunks as needed. Each actor is provided with a set of parametrized actions to perform precise edits to the document chunks, allowing for flexible and context-specific edits. The set of possible actions includes:

\begin{itemize}
    \item \textbf{Edit Chunk}: Modifies an existing chunk within a document by replacing content with updated text.
    \item \textbf{Add Chunk}: Creates a new chunk with specified content and adds it to the document.
    \item \textbf{Delete Chunk}: Removes an existing chunk from the document entirely.
\end{itemize}

The ReACT agent utilizes these reflections and iteratively generates a trajectory $t_0= {a_0, a_1, a_2 \cdot a_n}$ of edit actions to the document until the errors are resolved or the knowledge gaps are filled. This controlled editing process improves the accuracy of the RAG system by ensuring that the KB contains up-to-date and relevant information. After the completion of the actor runs, we generate the edit diffs for each document $d_i$ and pool them to generate the KB edit action  \( u = [d_i]_{i=1}^{|\mathcal{K}|}\) 

\section{Experimental Setup}
\label{sec:experimentsetup}

\subsection{Baseline}
While there has been a rich body of works in the area of prompt optimization, to the best of our knowledge, \method~is the first work targeting the feedback-driven textual Knowledge Base Editing problem. Therefore, to perform a holistic evaluation of \method~ we implement - \promptAgent-E, an extension of \promptAgent~ ~\cite{wang2023promptagent} for the KB editing task. \promptAgent~ formulates prompt optimization as a strategic planning problem using Monte Carlo Tree Search (MCTS). We have described our implementation on top of \promptAgent~ in appendix section \ref{sec:promptagent-e-baseline}
\subsection{Datasets}
Knowledge Base Editing can be useful for scenarios where the KB is either incomplete or incorrect. We evaluate on EVOR \cite{Su2024ARKSAR} which is a dataset of documentation for programming language \textit{Pony} which can be incomplete in details along with natural language to code questions in them. Similarly, it has two more datasets about custom versions of \textit{SciPy} and \textit{Tensorflow} with the original documentation of these libraries which must be adapted for these custom versions. We also use the also CLARK-News dataset \cite{erase} which is a natural language dataset of news articles of outdated factual information. We describe each dataset in detail in the appendix in \ref{sec:dataset_detail}. 

\subsection{System Configurations}
For our experiments, we set a maximum search depth of $3$, an expansion width of $3$, and a maximum of $5$ iterations. The UCT algorithm with an exploration constant of $2.5$ is used for expansion nodes. The parameters are chosen to balance between effective exploration and computational cost.
\\
We set up a generic RAG system that uses an embedding similarity for semantic retrieval. Additionally, in line with prior works like ~\citep{zhang2023repocoderrepositorylevelcodecompletion} for coding-related tasks, we use an iterative retrieval setup wherein we first generate a code using naive retrieval and then query the database again with both the question and the generated code to improve the quality of retrieval before generating the final result. 
We use \embeddingModel~ as the embedding model and use cosine similarity as a metric of embedding match for ranking.

\subsection{Metrics}
\textbf{Correctness}\quad We evaluate the correctness of the KB by evaluating it on a \textit{test set} queries on respective tasks. This separation of the test-train set of queries reduces the risk of contamination of the examples and falsified improvements in performance.  
We also define two metrics \textit{completeness} and \textit{coherence} to understand the quality of the edits made by \method. 

\textbf{Completeness}\quad A knowledge base should be \textit{complete} with respect to the task, that means it should contain all the information necessary to assist RAG system for task at hand.
Given the open-ended nature of tasks that typical RAG agents are designed for, it is hard to quantify a closed-form metric of \textit{completeness}. However, an ideal KB editing system should at least be able to incorporate external feedback well. To evaluate this we use the precision \textit{train set} to estimate the degree of expert feedback incorporated in the learned KB.

\textbf{Coherence}\quad
Given the semantic and textual nature of the Knowledge Base, it is important that the documents in the Knowledge base are coherent and consistent even after editing. This not only makes the document interpretable for human consumption, it also help reduce in-context noise during LLM inference, which has been shown to affect LLM performance~\citep{liu-etal-2024-lost}. 
To quantify the degree of coherence of the KB, we first calculate coherence scores for each edited document using G-Eval~\citep{Liu2023GEvalNE}.
We use the G-eval score to gauge the coherence of an edit made to a KB document to the document itself. And the mean of this document-level coherence over all the documents is defined as the coherence of an edited KB. 
\begin{table*}[t]
\centering
\resizebox{\textwidth}{!}{%
\begin{tabular}{@{}lccccclcccc@{}}
\toprule
& \multicolumn{4}{c}{Completeness (in \%)} & & \multicolumn{4}{c}{Coherence (1$\to$5, higher is better)} \\ \midrule
\multicolumn{1}{c}{Dataset} & Pony & SciPy & Tensorflow & CLARK-news & & Pony & SciPy & Tensorflow & CLARK-news \\ \midrule

\method~GPT-4Turbo  & 9.68 & 31.38 & 44.44 & 13.79 & & 4.6 & 4.30 & \textbf{4.0} & 1 \\
\method~GPT-4o  & 11.38 & 36.67 & 50.12 & 15.41 & & \textbf{4.67} & \textbf{4.67} & \textbf{4.0} & \textbf{2.33} \\
\method~GPT-4.1  & \textbf{13.45} & \textbf{41.46} & \textbf{52.24} & \textbf{18.62} & & 4.6 & 4.00 & 3.67 & 1 \\
\bottomrule
\end{tabular}%
}
\caption{Completeness and coherence comparison between \method~and baseline models across multiple datasets. Completeness is reported as accuracy percentages (higher is better), while coherence is measured on a scale of 1-5 (higher is better).}
\label{tab:coherence-and-completeness}
\end{table*}

\section{Results and Analysis}
\label{sec:results}
We evaluate \method~on three different OpenAI models \textit{GPT-4Turbo}, \textit{GPT-4o} and \textit{GPT-4.1}\footnote{\href{https://platform.openai.com/docs/models}{https://platform.openai.com/docs/models}} on three different configurations. Firstly, evaluating the performance on the Base KB, this is the initial state of the KB $s_0$ without any edits. We then make a series of edits on $s_0$ using \promptAgent-E and a series of edits by \method~ on KB state $s_0$.

We report our main results in table \ref{tab:baseline-comparision}. We observe the performance of the RAG system constantly improve with edits done by \method for all the three models.

\textbf{Quality of edits}\quad
As seen in Table \ref{tab:coherence-and-completeness}, \method~produces edits with a coherence score of 4 or higher. For KBs that need long-term maintenance (such as language and code documentation as seen in the Evor datasets), \method~makes more coherent edits compared to the baseline. This is especially true for long documents, as seen in the EVOR Pony dataset. We also note that the edits made by \promptAgent-E were made in incorrect documents leading to more noisy generations. 
\\
We also observe that \promptAgent-E added irrelevant section in example \ref{sec:edit_example} on the \textit{lineSearch} and \textit{norm\_ppf} functions in a document about sparse matrices. These edits were made because the document was retrieved for questions which had errors regarding these functions. These edits are irrelevant to the document and showcase reason for failure in \promptAgent-E is to make documents less coherent.

On the other hand, \method~makes more relevant edits to the document which are contained to the context of the document. This shows how \method~is able to maintain the coherence of the document in its edits. We demonstrate an complete example of edits made in appendix \ref{sec:edit_example}
\\
For a news-article-like dataset like CLARK-news with factual edits. Incoherency is naturally induced when the facts of the article change. In this dataset, coherence is sacrificed in bringing the facts of the article up to date, which is required to improve accuracy. 

We also evaluate performance on EVOR-Pony dataset using just a single \method~edit and greedy search in table \ref{tab:search_comparison}. In greedy search, we greedily pick the most rewarding node at a particular depth. We observe even though the completeness of the document is quite similar the edits are much less coherent and the generalize less than MCTS based search. 
\begin{table}[]
\resizebox{0.48\textwidth}{!}{%
\begin{tabular}{@{}lccc@{}}
\toprule
             & Single Edit & Greedy Search & MCTS  \\ \midrule
Correctness  & 36.14     & 41.34         & 45.62 \\
Completeness & 9.68      & 13.96         & 13.45 \\
Coherence    & 4         & 3.34          & 4.6   \\ \bottomrule
\end{tabular}%
}
\caption{Comparison between single edit, greedy search and MCTS, on EVOR Pony dataset with GPT-4.1. We used max width=3 and max depth=5.}
\label{tab:search_comparison}
\end{table}

\subsection{Generalization to Repository Level Migration}
\begin{table}[]
\centering
\resizebox{0.48\textwidth}{!}{%
\begin{tabular}{@{}lccc@{}}
\toprule
\multicolumn{2}{c}{} & \multicolumn{2}{c}{\textbf{Migration Efficacy (in \%)}} \\
\cmidrule(lr){3-4}
Model & Method & $\eta_{\text{minimal}}$ & $\eta_{\text{maximal}}$ \\
\midrule
\multirow{2}{*}{\textbf{GPT-4o}} & Base KB & 43.67\% & 18.12\% \\
                                 & \method & \textbf{46.14\%} & \textbf{23.02\%} \\
\midrule
\multirow{2}{*}{\textbf{GPT-4.1}} & Base KB & 58.81\% & 26.50\% \\
                                 & \method & \textbf{61.71\%} & \textbf{28.16\%} \\
\bottomrule
\end{tabular}%
}
\caption{Test set performance of \method~and Base KB on Migration Bench}
\label{tab:migration-bench}
\end{table}
To further test the efficacy of \method~, we apply \method~to a more complex real-world agentic scenario, namely repository-level code editing. We choose to work with the \citep{migrationbench} benchmark, which consists of migration of 5,102 real repositories from Java 8 to Java 21 along with a comprehensive evaluation harness for any system attempting this task. This benchmark presents unique challenges including a substantially larger action space requiring navigation of entire repository structures, more delayed feedback signals determined by compilation and test execution, and specialized knowledge requirements for Java version-specific APIs and migration patterns.

For the purpose of this study, we extend the \textbf{SDFeedback} agent introduced in \citep{migrationbench} to incorporate KB-guided editing, where the agent queries the knowledge base before making each edit decision. For the KB, we manually create a small knowledge base consisting of basic information relevant to Java migration, hereby referred to as the BaseKB for this setting. The BaseKB contains 47 entries covering common migration patterns such as API replacements, deprecated method alternatives, and frequently encountered compilation errors with their resolutions. We deliberately keep the BaseKB small and somewhat incomplete to better demonstrate \method's ability to identify and fill knowledge gaps.

We use the \textbf{Selected} subset of the benchmark as our test set, and we randomly sample 100 examples from the remaining \textbf{Full} subset of the benchmark, dividing them equally to create training and validation sets. We run \method~with the same MCTS settings as the other datasets. We run this experiment with two models, \textit{GPT-4o} and \textit{GPT-4.1}. We use the metrics introduced in \citep{migrationbench} to represent the success rate of the migration, namely minimal migration ($\eta_{minimal}$), which checks whether the Java version is correctly updated and whether the migrated repository passes all test cases, and maximal migration ($\eta_{maximal}$), which adds the additional criterion of all dependencies needing to be updated to their latest versions. The complete details of the experiment, including exact modifications made to \textbf{SDFeedback} and the manually created knowledge base, are present in Appendix Section \ref{sec:Migration_bench_setup}.

Table \ref{tab:migration-bench} shows the migration success rates for the Base knowledge base and the knowledge base after \method~was applied. We observe approximately 3\% absolute improvement in $\eta_{minimal}$ for both models over the Base knowledge base, showcasing the effectiveness of the edits in addressing both functional correctness and dependency management. These gains are particularly notable given the already strong baseline performance of the \textbf{SDFeedback} agent.

Figure \ref{fig:java-mig} shows that the edits made to the base knowledge base consist primarily of addition of new error types and patterns not present in BaseKB (15 new entries), expansion of existing entries with more detailed resolution steps including specific code examples (23 entries expanded), and corrections to incomplete guidance (9 entries corrected). This demonstrates how \method~adds missing information to the knowledge base while also expanding upon and correcting already existing information, allowing the agent to make better edits. Edits on a section of the knowledge base can be seen in Figure \ref{fig:java-mig}.
\\


\section{Conclusion}
We introduced \method, a novel framework for refining Knowledge Bases (knowledge bases) in Retrieval-Augmented Generation (RAG) systems using a multi-actor, centralized critic architecture. \method~ enables efficient knowledge base updates without retraining or altering model parameters by leveraging feedback-driven structured edits and textual gradients. 

Our approach achieved superior performance in preserving knowledge base in terms of coherence, consistency, and completeness, resulting in enhanced performance of RAG system.\\
\section*{Broader Impact}
\label{sec:broader-impact}
The deployment of Retrieval-Augmented Generation (RAG) systems in real-world applications such as AI-powered developer assistants, enterprise chatbots, and domain-specific information retrieval relies heavily on the correctness and reliability of the underlying knowledge bases (KBs). However, maintaining these KBs is a persistent bottleneck due to frequent changes in domain-specific knowledge and the lack of automated mechanisms for continuous KB refinement. Our proposed system, \textbf{STACKFEED}, addresses this challenge through a feedback-driven framework for automatic knowledge base editing that learns from expert or oracle feedback in real-time deployments.

Our design introduces a modular, actor-critic architecture that can be integrated into existing pipelines with minimal engineering overhead. By defining clear KB quality metrics—\textit{correctness}, \textit{coherence}, and \textit{completeness}. Our system provides actionable insights for both developers and auditors. This supports not only continuous improvement of deployed systems, but also regulatory compliance and human-in-the-loop oversight in high-stakes domains like healthcare, finance, and legal automation. By automating feedback incorporation into KBs, we reduce human maintenance cost, lower response errors in production RAG systems, and promote safer, more trustworthy deployments.

In industrial RAG based applications, post-deployment error-correction and maintenance is done through improving the quality of the retrieval system. We introduce another axis by enabling the optimization of the Knowledge base itself. It also paves the way for joint optimization of both the knowledge base and the retrieval mechanism, offering a more holistic and scalable solution to long-term system maintenance.

We hope this work paves the way for future industry adoption of learning-enabled infrastructure that maintains and improves itself in deployment, and encourages further exploration of editable memory systems as an alternative to end-to-end retraining for knowledge maintenance.

\section{Limitations and Future Work}
While this work presents a novel framework for feedback-driven knowledge base refinement, several limitations and corresponding avenues for future research should be acknowledged. In particular, one limitation of this work is the decoupling of the knowledge base optimization and retrieval. This work assumes that the retrieval component can correctly identify the right documents to retrieve. Failures originating from faulty retrieval cannot be holistically addressed by this system. A promising avenue for future work can be the joint optimization of both the knowledge base and the retrieval mechanism. Creating a unified framework that can decide whether to fix an error by editing a document or by tuning the retriever could offer a more holistic and scalable solution to long-term system maintenance.

\vspace{.2em}



\newpage
\bibliography{anthology}
\bibliographystyle{acl_natbib}

\clearpage
\appendix
\section{Appendix}
\subsection{Knowledge Base Editing as State Search}
\label{sec:KB_as_state_search}
In our problem setting, the Knowledge Base (\(\mathcal{K}\)) is defined as a collection of documents \(\mathcal{K} = \{D_i\}_{i=1}^n\). We assume each document consists of a number of chunks of text and can be represented as \( D_i = [ c_{ij} ] \). The state \( s \in \mathcal{S} \) of the system is represented by the current configuration of the KB, i.e., the content of all documents in \(\mathcal{K}\).

Given a query \( q_i \) and a set of retrieved documents \(\Gamma(q_i, \mathcal{K})\), the LLM \(\mathcal{M}\) generates an answer \( o_i \). When errors arise due to incomplete or incorrect information in the retrieved documents, our goal is to identify the optimal configuration of \(\mathcal{K}\) that improves the accuracy of the system’s responses. Thus, we define our state search problem as finding the best state \( s^* \) of the KB.

\textbf{State Space:} The state space \(\mathcal{S}\) encompasses all possible configurations of the KB. Each state \( s \) corresponds to a particular set of document contents, represented as: \(
s = \{ D_i \}_{i=1}^n,
\) where \( D_i \) denotes the content of document \( i \) and \(n\) is the number of documents in \(\mathcal{K}\). The state \( s \) captures the overall structure and content of the KB at any given point. We set \(s_0 = \mathcal{K}\).

\textbf{State Transition Function:} The state transition function \(\mathcal{T}(s, u)\) defines how the KB changes in response to the action \( u \) taken by the agent. Each action contains modifications to one or more documents within the KB, resulting in a new KB configuration. The state transition is formalized as: \(
s' = \mathcal{T}(s, u),
\) where \( s' \) is the new state of the KB after applying \( u \).

\textbf{Action Space:} The action space \(\mathcal{A}\) consists of list of diffs \( d_i \) corresponding to each document \( D_i \). Essentially, \( u = [d_i]_{i=1}^{|\mathcal{K}|}\).

\textbf{Environment:} We model the environment simply as a ``patch'' function, that takes the diff generated by the agent and patches the KB to produce the new state. 

\textbf{Optimization Objective:} Following Equation~\ref{eq:optim_objective}, our objective then is to find the optimal state \( s^* \) of the KB that maximizes the overall performance of the RAG system, as measured by a global reward function \( R \). The optimization problem is formulated as: \ignore{\priyanshu{Call back on reward function}}


\begin{gather}
R(s) = \frac{1}{\vert Q \vert}\sum_{q_i \in Q}g(\mathcal{M}(q_i, \Gamma(q_i, s)))
\label{eq:state_reward}
\end{gather}
\begin{gather}
s^* = \arg\max_{s \in \mathcal{S}} R(s) 
\label{eq:state-search-objective}
\end{gather}

where \( R(s) \) represents the cumulative reward of the KB state \( s \), reflecting its ability to support accurate and complete responses for a set of queries.

The reward function \( R(s) \) is derived from the expert feedback on the system’s generated answers and captures improvements in the KB.
By optimizing for \( s^* \), we ensure that the final state of the KB maximizes the overall accuracy and effectiveness of the RAG system, rather than focusing on an intermediate sequence of state transitions.

In summary, the state search formulation defines the problem of finding the optimal state \( s^* \) of the KB that maximizes the system’s performance. This approach enables us to make targeted, feedback-driven edits to the KB and achieve a refined, high-quality knowledge base that better supports accurate answer generation.

\subsection{Migration Bench}
\label{sec:Migration_bench_setup}
We share the KB used for Java migration 
\begin{promptlisting}[caption={MigrationBench KB}]
Java KB


---

## **`java.lang.UnsupportedClassVersionError`**

**Symptom:**

```
Exception in thread "main" java.lang.UnsupportedClassVersionError: 
com/example/App has been compiled by a more recent version of the Java Runtime
(class file version 65.0), this version of the Java Runtime only recognizes up to 61.0
```

**Cause:**

* The code is compiled with Java 21 but run with an older JVM (e.g., Java 17).
* Class file version 65.0 corresponds to Java 21.

**Fix:**

* Ensure the runtime JVM matches the compiler version:

  ```bash
  java -version
  javac -version
  ```
* Upgrade your runtime to Java 21.

---

## **`module not found: java.base`**

**Symptom:**

```
error: module not found: java.base
```

**Cause:**

* Misconfigured module path.
* Incorrectly set `--release` or `--module-path` flags.

**Fix:**

* Verify `JAVA_HOME` points to JDK 21.
* Use:

  ```bash
  javac --release 21 ...
  ```
* Ensure dependencies are on the module path (or use classpath if not modularized).

---

##  **`invalid source release: 21`**

**Symptom:**

```
error: invalid source release: 21
```

**Cause:**

* Older build tool (Maven, Gradle, Ant) that does not yet support Java 21.

**Fix:**

* Upgrade the build tool version:

  * **Maven:** $\geq$ 3.9.x with `maven-compiler-plugin` 3.11+
  * **Gradle:** $\geq$ 8.3
* Example Maven `pom.xml`:

  ```xml
  <plugin>
    <groupId>org.apache.maven.plugins</groupId>
    <artifactId>maven-compiler-plugin</artifactId>
    <version>3.11.0</version>
    <configuration>
      <release>21</release>
    </configuration>
  </plugin>
  ```

---

## **`Preview feature used without --enable-preview`**

**Symptom:**

```
error: patterns in switch are a preview feature and are disabled by default.
```

**Cause:**

* Code uses **preview features** in Java 21 (e.g., string templates, unnamed classes, pattern matching in switch).

**Fix:**

* Compile and run with preview enabled:

  ```bash
  javac --enable-preview --release 21 MyApp.java
  java --enable-preview MyApp
  ```
* Avoid preview features in production code.

---

## **Gradle Daemon / Toolchain Errors**

**Symptom:**

```
Could not target platform: 'Java SE 21' using tool chain: 'JDK 17 (17)'.
```

**Cause:**

* Gradle is using an older JVM despite Java 21 being installed.

**Fix:**

* Configure toolchain in `build.gradle`:

  ```groovy
  java {
      toolchain {
          languageVersion = JavaLanguageVersion.of(21)
      }
  }
  ```
* Ensure `JAVA_HOME` points to JDK 21.

---

## **`cannot find symbol` with Standard APIs**

**Symptom:**

```
error: cannot find symbol
  symbol:   class SequencedCollection
```

**Cause:**

* Using new Java 21 APIs (e.g., `SequencedCollection`, `VirtualThread`) but compiling with an older JDK.

**Fix:**

* Compile and run with JDK 21.
* Ensure IDE is configured with Java 21.

---

## **`Illegal reflective access` warnings**

**Symptom:**

```
WARNING: An illegal reflective access operation has occurred
```

**Cause:**

* Libraries using reflection to access internal JDK APIs, stricter in newer Java versions.

**Fix:**

* Update to latest versions of affected libraries.
* If unavoidable, use JVM args (not recommended for long-term):

  ```bash
  --add-opens java.base/java.lang=ALL-UNNAMED
  ```

---
\end{promptlisting}
\begin{figure*}
    \centering
    \includegraphics[width=0.8\textwidth]{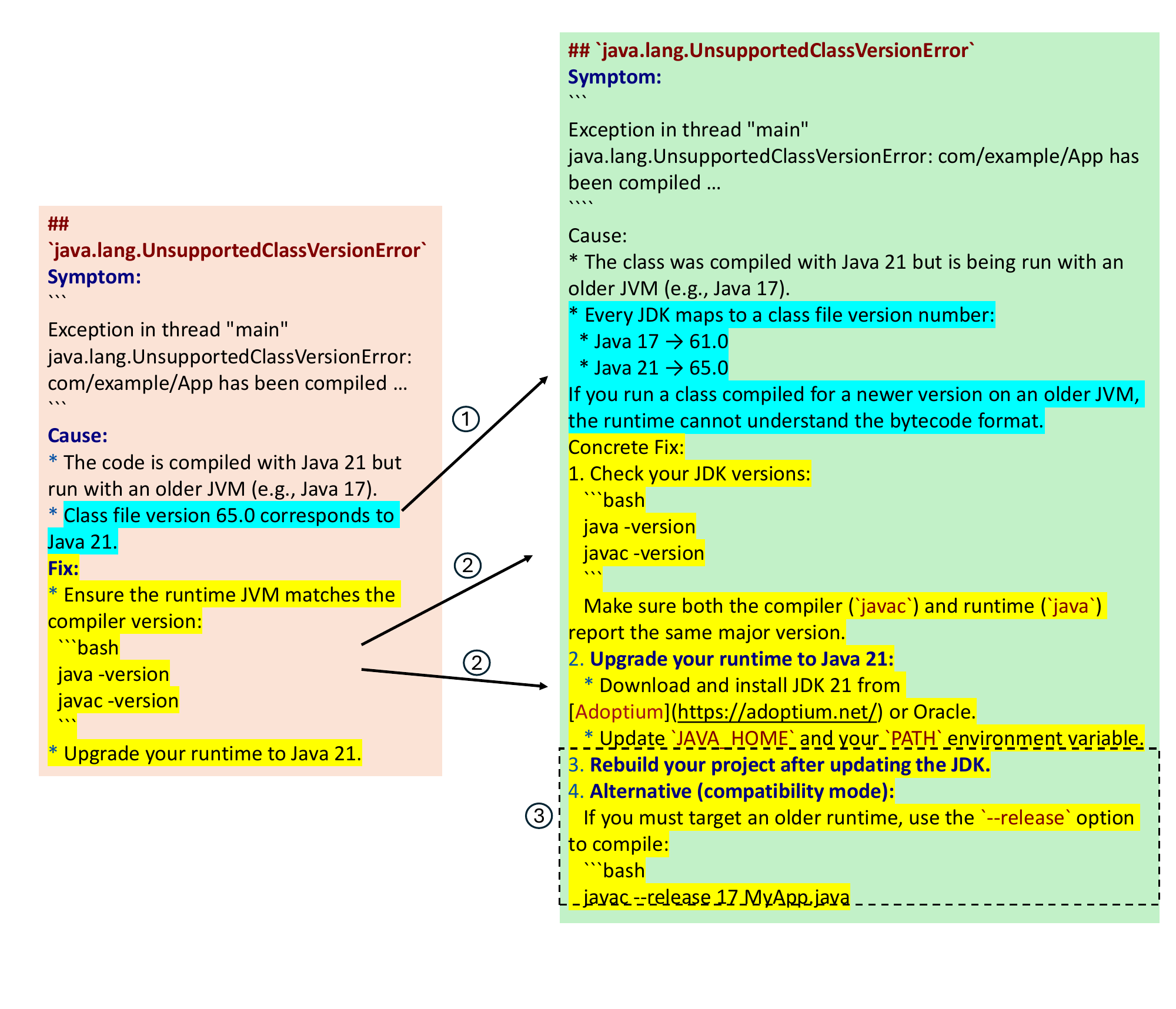}
    \caption{The above example showcases the edits made by \method.\textcircled{1} represents a more precise and structured state of information made from edits by \method. \textcircled{2} showcases a fix that was written more coherently and with added details for the agent by observations made from trajectory.\textcircled{3} showcase added information from train set on resolution.}
    \label{fig:java-mig}
\end{figure*}
\subsection{Example and Overview}
\label{sec:exampleoverview}
Figure~\ref{fig:example} illustrates our technique applied to the Pony ~\citep{Su2024ARKSAR}, where a knowledge base (KB) for the low-resource programming language Pony supports a natural language-to-code task. Due to Pony’s rarity, language models often generate code that fails to compile. To address this, we use the Pony compiler as an expert to provide feedback in the form of compile errors.

\circled{1} \emph{Evaluating the Knowledge Base State}:
We start with an initial KB, including documents like \texttt{builtin-array.md}. The system retrieves relevant documents based on the given task (e.g., counting non-inversions in an array) and generates a program, which is evaluated by the compiler, resulting in feedback (e.g., compile errors).





\circled{2} \emph{Centralized Critic Analysis}: For all errors, the critic analyzes why the error occurred. For instance, if the \texttt{apply} method in the \texttt{Array} class is partial and may raise an error, the critic suggests adding a \texttt{?} to handle potential failures.
Based on this reflection, the critic identifies which of the retrieved document is relevant for the error and provides a tailored reasoning for it.

\circled{3} \emph{Per-Document Actor}: For each document in the KB, the gradients associated to it are aggregated. This aggregate gradient is used as a signal by the Per-Document Actor, in this case, the actor for document \texttt{builtin-array.md} to make edits to the document.

\circled{4} \emph{Re-evaluation and MCTS Search}:
After edits are applied, the KB is re-evaluated, generating new feedback and a reward score. This score guides a Monte Carlo Tree Search (MCTS) to explore different states of the KB, iterating through steps \circled{1}-\circled{3} to progressively refine the KB and improve the system's overall performance.

\subsection{PromptAgent-E Baseline}
\label{sec:promptagent-e-baseline}

\promptAgent~\cite{wang2023promptagent} is a technique developed for optimizing a single prompt. \textbf{PromptAgent-E} extends this approach to knowledge base (KB) optimization by independently optimizing each document within the KB using a separate \promptAgent~instance. Unlike \method, \promptAgent-E operates as a collection of document-wise Independent Actor-Critic models~\cite{Foerster2017CounterfactualMP}.

\subsubsection{Algorithm Description}
The \promptAgent-E algorithm proceeds as follows:
\begin{enumerate}
    \item \textbf{Initial Evaluation:} Given a training set of queries, we first run the current KB to obtain retrievals, generations, and expert feedback for these generations. The same is done for the validation set.
    \item \textbf{Document-Level Dataset Creation:} The training set is then segmented into document-level training sets. Each document-level set comprises all queries (along with their corresponding retrievals, generations, and expert feedback) for which a specific document was retrieved. Similarly, the validation set is also split into document level validation sets.
    \item \textbf{Document Selection for Editing:} Given that KBs can be extensive, we restrict editing to documents that were retrieved for at least two queries in the training set. This ensures focusing on more relevant or frequently accessed documents.
    \item \textbf{Independent Optimization:} A separate \promptAgent~instance is then created and executed for each selected document independently. Each document-level \promptAgent~instance only accesses the queries, generations, and feedback pertinent to its assigned document.
    \item \textbf{KB Update:} After determining the optimal node (or prompt) for each document, these optimized nodes are integrated back into the KB to form a new, improved version.
\end{enumerate}

\subsection{Dataset}
\label{sec:dataset_detail}
Knowledge Base Editing can be useful for scenarios where the KB is 
\begin{enumerate}
    \item Incomplete: the knowledge bases misses some key artifacts responsible for answering the questions. In the Evor-Pony dataset, the documentation used lacks information on various aspects of the language like partial functions etc.
    \item Incorrect: the knowledge base in this case consists of some incorrect knowledge.
\end{enumerate}. 
We evaluate \method~ on 5 datasets spanning these different settings.
\subsubsection{Incomplete Knowledge Base}
 
\begin{table}
\centering
\resizebox{0.45\textwidth}{!}{%
\begin{tabular}{lcccc}
\hline
\multicolumn{1}{l}{Dataset} & Train & Eval & Test & Documents \\ \hline
Pony & 31 & 32 & 45 & 601 \\
ScipyM & 22 & 22 & 98 & 3921 \\
TensorflowM & 9 & 9 & 26 & 5859 \\
CLEvor News & 30 & 30 & 60 & 138 \\ \hline
\end{tabular}%
}
\caption{Dataset Statistics}
\label{tab:data-splits}
\end{table}
We adapt \textit{two} code generation datasets from Evor~\citep{Su2024ARKSAR}, namely \textbf{Evor-Pony}. The dataset consists of LeetCode problems and their solutions in low-resource languages Pony and Ring respectively. Each datapoint is supplemented with a corresponding language documentation, with execution accuracy as the success metric and execution failures as feedback to the system. Given that these languages don't appear prominently in LLM pre-training data, the performance of code generation RAG agents on these datasets depends significantly on the quality of the Knowledge Base. However, given that these languages have smaller communities, their documentation isn't as well maintained and often lacks critical information. \ignore{\priyanshu{Refer to example figure to motivate}}. For the purpose of evaluation on these datasets, we split them into train, eval, and test splits as specified in Table ~\ref{tab:data-splits}. To ensure that we have a good representation of failure cases during training, we first execute the RAG pipeline on the entire dataset and divide the failures at random in a 1:1:2 ratio for train, eval, and test respectively. All the datapoints with successful execution matches are put in the test split. We  use the compiler feedback from the executions as the expert feedback to the \method~system. 
\subsubsection{Incorrect Knowledge Base}
For evaluating under this setting, we leverage the \textbf{Evor-ScipyM} and \textbf{Evor-TensorflowM} datasets from Evor and the \textbf{CLARK-news} dataset from Erase~\citep{erase}. 
The Evor datasets consist of data science problems sourced from the DS-1000 dataset ~\citep{Lai2022DS1000}, which are to be solved by artificially perturbed versions of scipy and tensorflow libraries respectively, while referring to the original unperturbed documentation. Similar to Pony and Ring,  we use the execution accuracy on a test bench as a success metric and use compiler outcomes as expert feedback. We also follow a similar approach for data splitting.

While fact retrieval is one of the most popular use cases of RAG systems, evolving nature of information requires us to keep the knowledge bases up to date. To simulate these dynamic factual knowledge updates we use the CLARK-news dataset from Erase~\citep{erase} which contains questions and their respective answers extracted from Wikidata at different timestamps. Each timestamp is characterized by a set of articles that were added in the data at that time. For our evaluation, we pool all the questions whose answers changed for the \textit{first} time at a given timestamp and split them across train, eval and test splits in a 1:1:2 ratio (Table \ref{tab:data-splits}).

\subsection{Example of Automatic Edits in Evor-Scipy}
\label{sec:edit_example}
This is a case in which both PromptAgent-E and \method~ opt to append sections to the end of a document about \emph{sparse matrices} in the modified version of SciPy provided in the Evor Dataset.

Demonstrably, the baseline edits add unnecessary information from the \emph{newScience.algorithm} and \emph{newScience.Distribution} modules to the document. The document is about \emph{sparse matrices} so the addition of information about \emph{lineSearch} and \emph{norm\_ppf} is not appropriate for this document and it is causing the document to become incoherent

On the other hand, the edits made by \method~are relevant to sparse matrices and keep the document coherent.
\begin{lstlisting}[language=MyDiff,caption={PromptAgent-E Edits (Only showing the added sections)}]
+# newScience.algorithm Module

+## Functions

* lineSearch(func, grad, initial_point, direction, **kwargs)
    Perform a line search to find the step size (alpha)
    that satisfies the strong Wolfe conditions.

    Parameters:
    - func : callable
        The objective function to be minimized.
    - grad : callable
        The gradient of the objective function.
    - initial_point : ndarray
        The starting point for the line search. 
        Must be provided as a NumPy array.
  . . .

# newScience.distribution Module

## Functions

* norm_ppf(p_values)
Convert p-values to z-scores using the standard normal
distribution. This function is also known as the inverse
of the cumulative distribution function (CDF) for the
standard normal distribution.

Parameters:
- p_values : ndarray
    An array of p-values to be converted.

Returns:
- z_scores : ndarray
    The corresponding z-scores for the given p-values.

Example:
```python
import newScience.distribution as nsd
import numpy as np
p_values = np.array([0.05, 0.025, 0.975])
z_scores = nsd.norm_ppf(p_values)
print(z_scores)
\end{lstlisting}

\begin{lstlisting}[language=MyDiff,caption={STACKFEED Edits (Only showing added sections)}]
### Custom Sparse Matrix Classes

When working with custom implementations of sparse matrices, 
it is crucial to understand that the attributes and methods 
described in this documentation are specific to the 
`scipy.sparse.csr_matrix` class. 
Custom sparse matrix classes may not have the same interface or 
support the same operations.

#### Disclaimer on Custom Implementations

Developers should not assume that custom sparse matrix classes 
will have attributes like 'nnz' or methods such as 'nonzero'. 
Instead, they should refer to the documentation or 
implementation of the custom class to understand the available
attributes and methods.

##### Example of Incorrect Assumption

```python
# Incorrect assumption that a custom SparseMatrix class has 
'nnz' attribute
sparse_matrix = SparseMatrix(...)  # Custom sparse matrix class
if sparse_matrix.nnz == 0:
    print("The sparse matrix is empty.")
# This may raise AttributeError if 'nnz' is not an attribute of
SparseMatrix
Example of Correct Usage
python
Copy
Edit
# Correct approach using methods supported by the custom class
sparse_matrix = SparseMatrix(...)  # Custom sparse matrix class
if sparse_matrix.count_nonzero() == 0:
    print("The sparse matrix is empty.")
# Alternatively, convert to a dense array and use numpy's 
count_nonzero
dense_array = sparse_matrix.toarray()
if np.count_nonzero(dense_array) == 0:
    print("The sparse matrix is empty.")
\end{lstlisting}

\subsection{Prompts used in \texorpdfstring{\method}{F}}

\begin{promptlisting}[caption={Generalization Stage Prompt},label={fig:generalization-stage-prompt}]
"""
There exists a Language Model based software named CodeRAG that automatically does the following task for a developer:
{task} - {task_desc}

CodeRAG uses a knowledge base to perform this task:
{kb_desc}

A developer used CodeRAG to perform the task on multiple files, and CodeRAG made some errors on them.

Here is one knowledge base file that was involved in these errors:
"""
for i, file in enumerate(kb_files):
    prompt += f"""
File {i+1}:
id: {file['id']}
content: \n<file>\n{file['content']}\n</file>\n"""
    if "special_notes" in file and file["special_notes"] != "":
        prompt += f"""\nspecial_notes: {file['special_notes']}"""

"""
The following are the reflections on the errors made by CodeRAG:
{reflections_str}

The reflections show the relationship of the file with the errors made by CodeRAG.
If the file is named "None," it means the information about the error on which the reflection is based does not fully fit any knowledge base file.

Your task is to use the reflections on the errors made by CodeRAG and provide a generalization on the issues with the file and how it can be improved to prevent the errors.

You should mention common issues found in the reflections and provide a plan for improving the knowledge base files to prevent future errors. Use the reflections to suggest additions or changes in the file, explaining what new content should be added to prevent errors. Before suggesting your plan, give context on the errors using code snippets and other relevant information from the reflections.

You have a scratchpad to reason and plan your generalization. Your scratchpad is for your use only and will not be shared with anyone else.
The scratchpad is represented by the <scratchpad></scratchpad> tags.

Your generalization should follow this format:
<scratchpad>
The contents of the scratchpad
</scratchpad>
<generalization>
Your generalization for this file
</generalization>

You must provide the filled-out scratchpad and generalization in the above format.

General guidelines:
1. Carefully analyze the reflections to understand the errors CodeRAG is making.
2. "None" is a special file, representing that to fix the error, the information should be in a new file.
"""
\end{promptlisting}

\begin{promptlisting}[caption={Selection Stage Prompt},label={fig:generalization-stage-prompt}]
""" 
Your task is to reflect upon the errors made by CodeRAG based on the user feedback and provide a reflection on the role of the knowledge base files in the making of those errors.

Your reflection should be very specific to the knowledge base files as these reflections will be used to improve the knowledge base files to prevent such errors in the future. 

There may be other causes for the error, but you should only focus on whether the knowledge base files could have prevented the error. 

You should also provide a way for improving the knowledge base files to prevent the error from happening again. 

You should try and see if there is any error in the information provided by the knowledge base or if the knowledge base is missing some information that could have prevented the error.

You also have to figure out if the file should be edited or not. That you do through the needs_editing flag.

You have a scratchpad in which you can reason and plan your reflection. Your scratchpad is for your use only and will not be shared with anyone else. This scratchpad is represented by the <scratchpad> tags.

Your output should be in the following format:

<scratchpad>
The contents of the scratchpad
</scratchpad>

<reflection>
<File 1>
File: Name of the first file
needs_editing: True/False
Reflection: The reflection for this file
</File 1>

<File 2>
File: Name of the second file
needs_editing: True/False
Reflection: The reflection for this file
</File 2>
...
</reflection>

You have to provide the filled-out scratchpad and the reflection in the above-described formats. You have to reflect on all the files that were extracted for the code file.

Here are some general guidelines to follow:
1. You should first analyze the question, the test bench, the feedback, and the output to understand the error made by CodeRAG.
2. Then you should carefully analyze the knowledge base files to see if the theme and the contents of any knowledge base file are relevant to the error. Particularly, you should look out for files that have a factual error related to the error or are missing some information which should have been in the file according to the theme of the file.
    a. Read the content of the file and understand the theme of the file. The theme of this file is of course based on the file ID and the content of the file but you should also consider its positioning in the knowledge base. That means you should consider the other files that were extracted for the code file and see how this file fits in with them. For example, if the file is a very basic general guide to the task with other files providing more detailed information, then it would make sense for this file to not have detailed information about specific cases.
    b. See if the file has any information related to the error. Check for relevant keywords and how the file might have biased the language model to make the error.
    c. If the file has information related to the error, see if the information is correct and complete. If the information is incorrect or incomplete, the file is responsible for the error.
    d. If it doesn't have information related to the error, check if it makes sense for the file to have information related to the error. If it doesn't make sense, the file is not responsible for the error. When deciding this, check whether the information would be better suited in any of the other knowledge base files. If the missing info fits better in another file, then deem this file to not be responsible for the error as the missing content can be better placed in the other file.
    e. If the file is responsible for the error, explain the error in your reflection and set the needs_editing flag to True. And if the file is not responsible for the error, set the needs_editing flag to False.
3. If none of the files have any error or if you think the content for the error should be in a new file, put a file with the name "None" in your reflection and for its reflection, describe the error and mention why it is not due to the knowledge base files. For the "None" file, the needs_editing flag should always be set to True. The "None" file should be placed as File n+1 where n is the number of files extracted for the code file.
4. Choose the least number of files for editing, we want to change as few files as we can for any error. For example, if we have 5 knowledge base files, unless very extreme cases, we wouldn't want to set the needs_editing flag as True on more than 2 files. Figure out what the most relevant files for the error are and focus on them.
5. When you choose to edit multiple files, you should make sure that their involvements in the error are distinct and not overlapping. If they are overlapping, think about whether changing one file would be enough to fix the error.
"""
\end{promptlisting}


\begin{promptlisting}[caption={Reflection Stage Prompt}]

""" 
There exists a Language Model based software named CodeRAG that automatically does the following task for a developer:
{test_bench_code}

The test bench code gives a code where a function must be inserted and then it is tested with some

test cases.

CodeRAG then outputted the following code to answer the question:

if task_desc != "":
    prompt += f"""
{task} - {task_desc}
"""
else:
    prompt += f"""
{task}
"""

prompt += f"""

The developer used CodeRAG for a question. The question is as follows:
{query}

In the question, the developer provided the following test bench code:
{test_bench_code}

The test bench code gives a code where a function must be inserted and then it is 

tested with some test cases.

CodeRAG then outputted the following code to answer the question:
{output_code}

Based on the above output, the developer gave the following feedback to CodeRAG:
{feedback}

CodeRAG uses a knowledge base to do this task
{kb_desc}

The following files were extracted for this particular code file (the content of 

each file is surrounded in <file></file> tags):
"""
for i, instruction in enumerate(instructions):
    prompt += f"""
File {i+1}:
id: "{instruction['id']}"
content: \n<file>\n{instruction['content']}\n</file>\n
"""
    if "special_notes" in instruction and instruction["special_notes"] != "":
        prompt += f"""\nspecial_notes: {instruction['special_notes']}"""

prompt += """
Your task is to reflect upon the errors made by CodeRAG based on the user feedback. 
You have to explain in detail the error made by CodeRAG. The reflection should be 

very specific to the question, the output code and the feedback.
You should start by explaining the question that CodeRAG was asked to solve before talking about the error.

Your reflection should have relevant code snippets from the output 

code which have errors and what should be done to fix them.
You should also add a small code example to demonstrate the error and potential methods to fix it.

You can talk about multiple different methods here to address the error.

You have a scratchpad in which you can reason and plan your reflection.

Your scratchpad is for your use only and will not be shared with anyone else.

Your reflection should be in the following format:
<scratchpad>
The contents of the scratchpad
</scratchpad>
<reflection>
Your reflection
</reflection>
"""
"""
    
\end{promptlisting}

\end{document}